\documentclass{article}
\usepackage{ijcai19}

\usepackage{times}
\usepackage{soul}
\usepackage{url}
\usepackage[hidelinks]{hyperref}
\usepackage[utf8]{inputenc}
\usepackage[small]{caption}
\usepackage{subfigure}
\usepackage{graphicx}
\usepackage{amsmath,amssymb,amsfonts}
\usepackage{algorithm}
\usepackage{algorithmic}
\usepackage{mathtools}
\usepackage{xcolor}
\usepackage{booktabs}

\definecolor{bostonuniversityred}{rgb}{0.8, 0.0, 0.0}
\definecolor{royalazure}{rgb}{0.0, 0.22, 0.66}
\definecolor{ao(english)}{rgb}{0.0, 0.5, 0.0}

\title{Multi-hop Federated Private Data Augmentation with Sample Compression}

\author{
	Eunjeong Jeong$^1$\and %\footnote{contact info}
	Seungeun Oh$^1$\and
	Jihong Park$^2$\and
	Hyesung Kim$^1$\and\\
	Mehdi Bennis$^2$\And
	Seong-Lyun Kim$^1$\\
	\affiliations
	$^1$Electrical and Electronic Engineering, Yonsei University, Seoul, Korea\\
	$^2$Centre for Wireless Communications, University of Oulu, Finland\\
	\emails
	\{ejjeong, seoh, hskim, slkim\}@ramo.yonsei.ac.kr,
	\{jihong.park, mehdi.bennis\}@oulu.fi
}

\begin{document}
	
\maketitle
	
	\begin{abstract}

		On-device machine learning (ML) has brought about the accessibility to a tremendous amount of data from the users while keeping their local data private instead of storing it in a central entity. % On-device machine learning brings about multiple benefits including data privacy, faster processing, and offline accessibility. Specifically, privacy preservation is a strong point in that consumers in modern society concern the security of their sensitive personal data. 
		% In local machine learning process, a device can avoid disclosing its sensitive local data. At the same time, by cooperating with other devices, each one acquires a well-trained model that cannot be achieved by oneself.
		However, for privacy guarantee, it is inevitable at each device to compensate for the quality of data or learning performance, especially when it has a non-IID training dataset. %Preserving privacy by attaching dummy samples accompanies additional communication payload. Moreover, sample distortion before transmission can eventually damage the performance of local training.
		In this paper, we propose a data augmentation framework using a generative model: \emph{multi-hop federated augmentation with sample compression (MultFAug)}. %For the data augmentation, a device uploads its seed samples via a relay network and expects the server to train a generative model taking account of those samples.
		A multi-hop protocol speeds up the end-to-end over-the-air transmission of seed samples by enhancing the transport capacity. The relaying devices guarantee stronger privacy preservation as well since the origin of each seed sample is hidden in those participants. % and enhanced privacy guarantee of each device.
		%The relaying transmission also benefits the devices by  obfuscating the source of the samples as they are tossed to and aggregated on relaying devices.
		For further privatization on the individual sample level, the devices compress their data samples.
		The devices sparsify their data samples prior to transmissions to reduce the sample size, which impacts the communication payload. This preprocessing also strengthens the privacy of each sample, which corresponds to the input perturbation for preserving sample privacy.	 
		%On the other hand, the sample sparsification let the communication more efficient by reducing the size of seed samples. 	
		%\tblue{We show the influence of the number of hops and sparsification rate to uplink latency and privacy preservation. Before the evaluation, we suggest indices of label and sample privacy guarantee.} 
		The numerical evaluations show that the proposed framework significantly improves privacy guarantee, transmission delay, and local training performance with adjustment to the number of hops and compression rate.

	\end{abstract}

	\section{Introduction}

	% On-device ML: Fueled by the recent advances in computing power
	% Problem: Lack of data samples
	% - Way 1. Data downloading: communication bottleneck
	% - Way 2 [NeurIPS]. FAug exploiting cloud server's computing power and fast connection to the Internet: collectively train a generator to be downloaded, and locally augment data samples
	
	% Focus: During the seed sample collection, the impact of (1) multi-hop communication and (2) sample compression on FAug, in terms of communication efficiency, privacy guarantee, and FAug performance (quality of the generating samples, its impact on on-device learning)
	
	% Benefits of (1) multi-hop communication:
	% 1) Path loss reduction under limited Tx power
	% 2) Higher label privacy guarantee
	
	% Benefits of (2) sample compression:
	% 1) Payload size reduction
	% 2) Higher sample privacy guarantee

	Enhanced accessibility to big data has enabled machine learning(ML) and its further application as exemplified by on-device machine learning~\cite{Park:2018aa,Brendan17,Jeong18}. Individual devices are able to train a local model based on their private data, such as e-health medical records. %Recent advances in computing hardware has powered on-device machine learning (ML)~\cite{Park:2018aa,Brendan17,Jeong18}, wherein each device trains a local ML model using its user-generated private data (e.g., e-Health medical records).
	Under limited communication resources and privacy constraints, it is a key challenge to achieve a ML model with high accuracy in real-life situations. %One key challenge in on-device ML is to achieve high test accuracy subject to limited communication resources and privacy requirements.
	Furthermore, each device is often scarce in samples and biased, i.e., non-IID across devices, which hampers generalization of the local model against unseen data samples. %Indeed, the user-generated training data often lacks samples while being severely biased, i.e., non-identically and independently distributed (non-IID) across devices, making the local model unable to generalize against unseen data samples~\cite{Park:2018aa,Brendan17}.
	
	\begin{figure} [!t]
		\centering
		\subfigure[MultFAug.]{\includegraphics[width=\columnwidth]{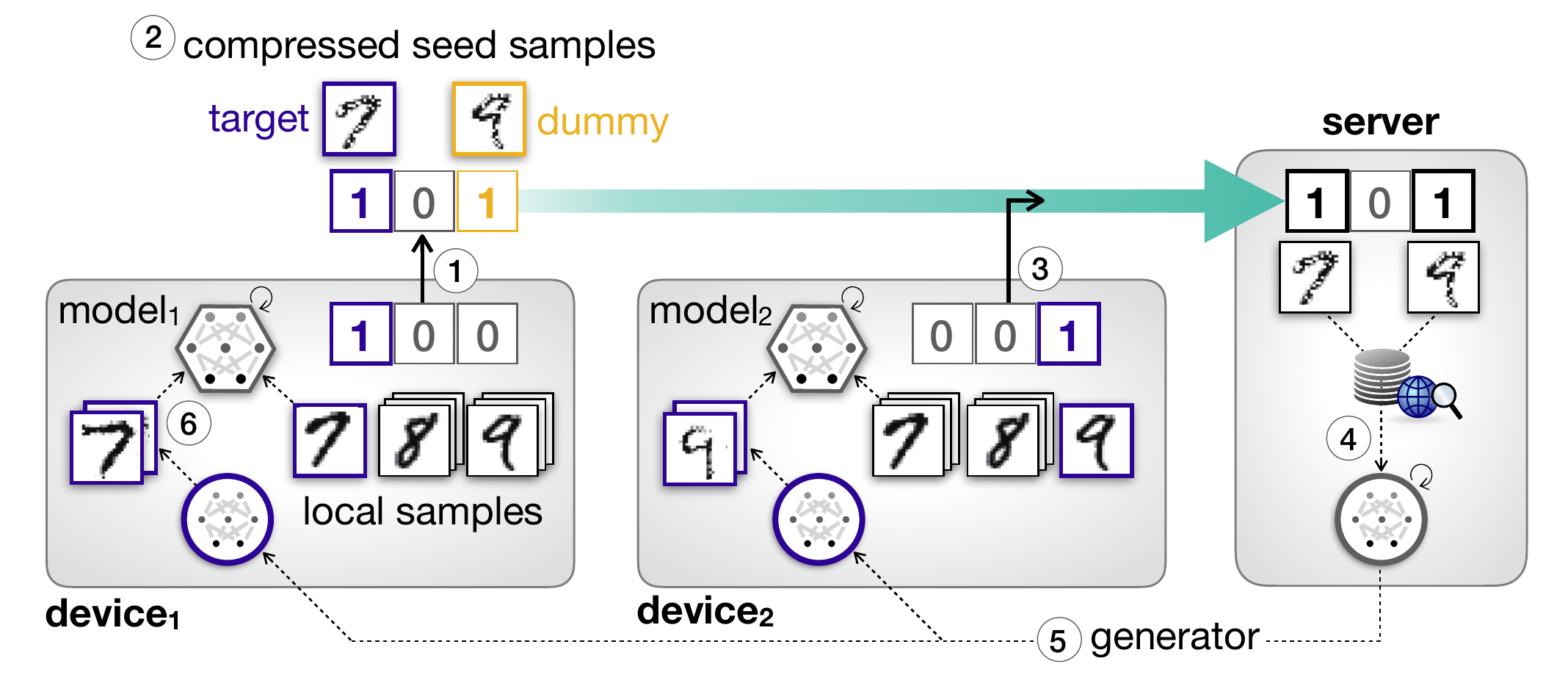} }
		\subfigure[Single-hop FAug without sample compression.]{\includegraphics[width=\columnwidth]{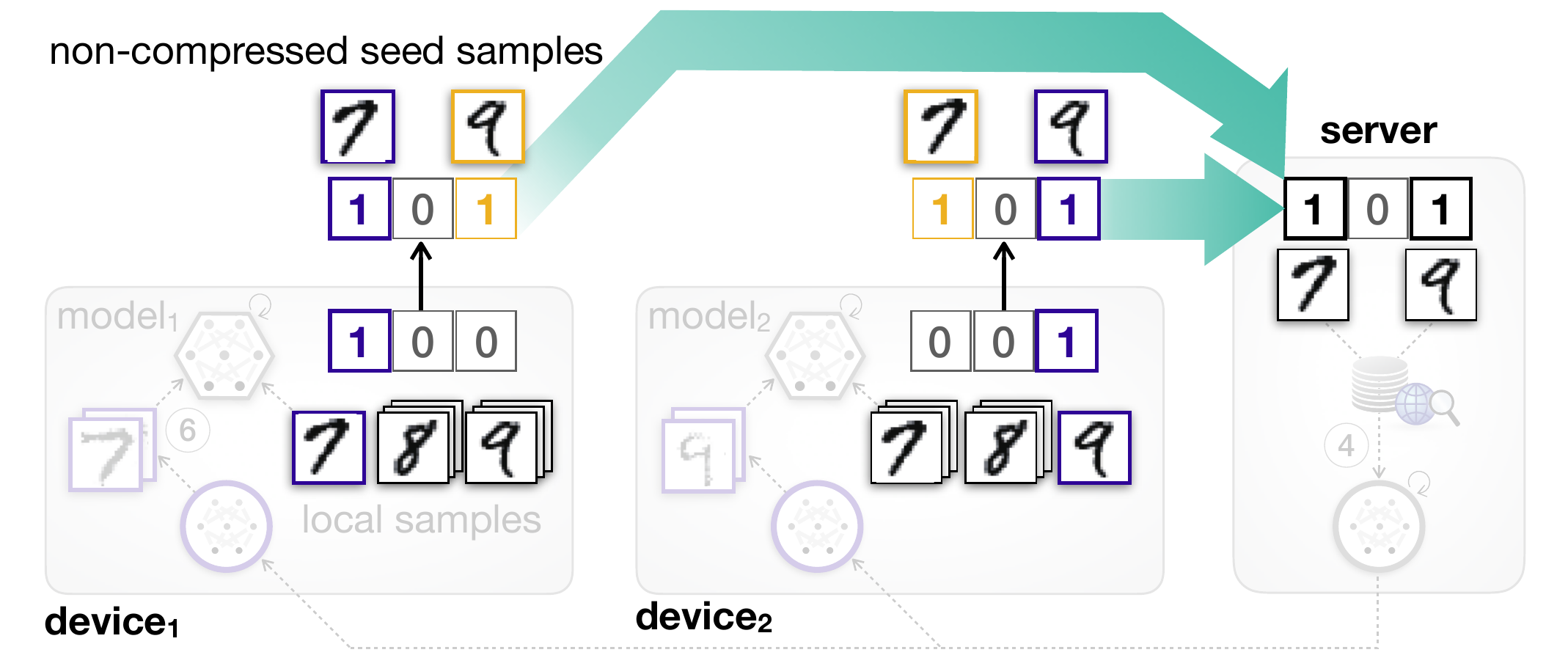}} \vskip-5pt
		\caption{\small Comparison between (a) \emph{multi-hop federated augmentation with sample compression (MultFAug)} and (b) single-hop FAug without sample compression, for $2$ devices associated with a~server.} \vskip-10pt
		\label{Fig_overview}
	\end{figure}

	To cope with this, we proposed \emph{federated augmentation (FAug)} in our preceding work~\cite{Jeong18}, in which the devices collectively train and share a data sample generator. The associated edge server builds and trains a generator from a conditional generative adversarial network model(cGAN)~\cite{Mirza14}, enabling each device to augment data samples by downloading the generator.
	%A straightforward solution is to exchange data samples across devices, which may unfortunately consume huge communication overhead while violating data privacy~\cite{Lu:19}. Alternatively, in our preceding work we proposed \emph{federated augmentation (FAug)}~\cite{Jeong18}, in which devices participate in training a sample generator (e.g., a conditional generative adversarial network (cGAN) model~\cite{Mirza14}) at their associated edge server, and by downloading the trained generator each device can locally augment data samples.
	The generator training at the server requires only a few seed samples collected from those devices. %This is viable by collecting few seed samples at the edge server to train the generator.
	Thanks to these small number of seed samples, the scheme relieves communication overhead as well as privacy leakage compared to exchanging data samples directly across devices. However, it is inevitable to compensate privacy for uplink capacity. In other words, each device should upload more seed samples so as to reach strong privacy preservation by hiding their weaknesses in the large volume of samples.
	
	In this paper, we propose a \emph{multi-hop federated augmentation with sample compression (MultFAug)} to improve both privacy guarantee and communication efficiency of on-device learning. Extending from the seed sample collection of FAug (see Figure~1(b)), multi-hop communication allows the devices to preserve privacy as they can hide their data distribution in the crowd. It also reduces the latency through succeeding short-distance transmissions (see Figure~1(a)).
	%In this paper, focusing on the seed sample collection procedure of FAug (see Figure~1(b)), we propose a \emph{multi-hop federated augmentation with sample compression (MultFAug)} collecting compressed seed samples through multiple hops (see Figure~1(a)). This improves the communication efficiency and privacy guarantee of on-device learning.
	Namely, compressing seed samples reduces communication payload sizes while preserving more \emph{sample~privacy}, at the cost of compromising the quality of augmented samples. On the other hand, multi-hop communication allows each device to hide its data distribution over data labels, denoted as \emph{label privacy}, in a communication-efficient way. These benefits are exemplified in Figure~1(a) describing the MultFAug operations as follows.

	\begin{enumerate}
		\item Out of $3$ labels (digits $\mathsf{7}$, $\mathsf{8}$, $\mathsf{9}$), $\textsf{device}_1$ lacks the samples of the \emph{target label} $\mathsf{7}$. To preserve the label privacy by hiding its private sample distribution information (SDI) $[1,0,0]$, $\textsf{device}_1$ inserts a \emph{dummy label} indicator into $\mathsf{9}$, yielding the public SDI $[1,0,1]$.
		
		\item $\textsf{device}_1$ compresses and appends two seed samples in $\mathsf{7}$ and $\mathsf{9}$ to its public SDI, and then transmit them to the next hop $\textsf{device}_2$~(see Algorithm 1).
		
		\item $\textsf{device}_2$ has the target label $\mathsf{9}$. Its private SDI $[0,0,1]$ can be hidden within the public SDI $[1,0,1]$ of $\textsf{device}_1$. Therefore, $\textsf{device}_2$ only forwards its received seed samples to the server.
		
		\item After collection, the server first oversamples the seed samples (e.g., via Google's image search for visual data) and then trains a generator model~(see Algorithm 2).
		
		\item Each device downloads the trained generator, and thereby locally augments the samples in the target labels for training its on-device ML model.
	\end{enumerate}

	Compared to the original FAug, we numerically validate that the proposed MultFAug achieves the same test accuracy of the on-device ML model, with shorter communication latency while preserving more data privacy.
	
	\vskip 5pt
	\noindent \textbf{Contribution}.\quad
	Our contributions in this paper consist of two main idea as follow:
	\begin{itemize}
		\item \emph{multi-hop communication}. We propose a data augmentation framework that improves communication efficiency and privacy preservation by adopting wireless transmission over multi-hop protocol.
		\item \emph{data compression}. The deletion of randomly picked bits in data samples leads to less communication overhead and a stronger privacy guarantee for each data sample.
	\end{itemize}

	\vskip 2pt
	\noindent \textbf{Related works}.\quad
	%\tred{discuss more related work to achieve a complete literature review.}
	To address the lack of training data samples, it is common to locally oversample the original data, e.g., via rotating and masking for visual data~\cite{Park:2018aa}, which may however struggle with locally biased samples. Exchanging data samples across different devices/servers is free from this problem, yet may incur huge communication overhead and/or violate data privacy~\cite{Balcan12}. Combination with federated learning alleviates the privacy violation problem~\cite{HybridFL}, but makes a strong assumption that some devices are insensitive to privacy and willingly provide their data samples as proxy data. On the other hand, global proxy data can be either collected or gathered at the edge server~\cite{Zhang17,Huang18}, yet it costs an additional effort to acquire such a dataset.
	Alternatively, generating synthetic data samples via generative adversarial networks (GANs) is a compelling data augmentation solution that guarantees data privacy while avoiding data exchanges~\cite{Sixt18,Zhu17}. These works however focus only on data sample privacy while neglecting SDI privacy and the impact on communication efficiency. In a similar manner, \cite{Yonetani19} suggested a system wherein users with non-IID dataset train discriminators independently. However, these local discriminators reveal the data distribution of each user since they reflect the non-IIDness of the individuals. By contrast, our preceding work~\cite{Jeong18} only considers SDI privacy and communication efficiency, without incorporating data sample privacy guarantees. In this work, we fill such a gap by incorporating both SDI and data sample privacy, and investigate the effectiveness of multi-hop communications for preserving privacy in a more communication-efficient way.
	
	%The rest of the paper is organized as follows. In Section II we describe our network model and framework of FAug, which is a baseline scheme for our MultFAug. The evaluation of latency and privacy is also introduced. In Section III we suggest MultFAug with sample sparsification and provide its framework with generalized evaluation of latency and privacy. The numerical results are shown under consideration of wireless channel model in detail in Section IV, and conclusion is drawn in Section V.

	\section{Single-hop Federated Augmentation Without Sample Compression}

	In this section, we describe a baseline model FAug wherein every device is directly connected to the server. This will be extended in the next section to a multi-hop version of FAug, MultFAug. 
	
	With this in mind, we consider a server and a group of edge devices in a network. The devices are denoted by $\mathcal{D}=\{1,2,...,N\}$ with $|\mathcal{D}|=N$, and their locations are randomly selected following uniform random distribution.
	%When linking the devices into a single route, a device chooses its nearest neighbor preferentially. To avoid an endless stagnation of samples, the candidate neighbors that have already been selected as relaying nodes are excluded from the selection. A collection of maps with different number of devices and hops is exemplified in Figure~\ref{fig_topology}.
	
	The devices employs time division multiple access (TDMA), in which each device fully occupies the bandwidth allocated along its route. The total bandwidth is equally divided into the number of routes and therefore equal range of bandwidth is assigned to each route.
	%The bandwidth is equally allocated to each route, which is the same as the total bandwidth divided into the number of routes in the system.	
	The devices transmit signals over the air experiencing path loss and fading. Under the system bandwidth $B$ and noise power spectral density $N_0$, $\textsf{device}_i$ sends samples with the instantaneous data rate $B \log_2 (1+ (g_i(t)P_i d_i^{-\alpha} / N_0B) ) $, where $g_i(t)$ is the fading coefficient, $\alpha$ is the path loss exponent, and $d_i$ denotes the distance between $\textsf{device}_i$ and its direct destination.
	When a device transmits packets at a lower data rate than its channel capacity at the moment, it retransmits with infinite number of attempts. %In case of transmitting at a lower data rate than the current channel capacity, devices retransmit with unlimited number of attempts.
	The transmission latency from $\textsf{device}_i$ to the server, referred to as $T_i$, is calculated as the number of time slots required to transmit all of the seed samples.

	%The device needs supplement of those missing data samples by data augmentation. As a solution for saving communication overhead in data augmentation, federated augmentation (FAug) is suggested in \cite{Jeong18}.
	%Seed samples show which labels are insufficient in the device's training dataset.

	Each device aims to train its learning model to classify data from its training set, which consists of image data samples and class labels corresponding to each sample. These class labels are discrete values tagged to a group of samples with common property.
	Let us assume that each device holds a non-IID training dataset; some of its class labels are deficient in data samples -- referred to as \textit{target labels}. 	
	In FAug, a device reports its missing labels to compensate for the target labels by uploading the seed samples to the server. When a device picks up seed samples, it must include target samples that the device wants to replenish.
	As a first step, each device uploads its seed samples directly to the server.
	To avoid unnecessarily large communication payload in case of sending too many samples with the same labels, the device restricts the number of seed samples per label to a predetermined upper bound value $b$. 
	The server oversamples these samples to train a conditional generative adversarial network (cGAN), one of the GAN variations in which extra information is additionally given as a condition for training both generator and discriminator~\cite{Mirza14}. When the training process ends, the devices receive the generator from the server. With the generative model, each device replenishes the target labels to achieve an IID training dataset.
	
	%GAN consists of two network models trained competitively: a generator and a discriminator. A generator $G$ is trained to synthesize realistic samples, while a discriminator $D$ is on the adversarial side trying to classify whether the data is original or made by the generator. To be specific, $G$ builds a mapping function from the prior random noise $\boldsymbol{z}$ to mimic the original sample $\boldsymbol{x}$ by learning the distribution of $\boldsymbol{x}$. Meanwhile, $D$ compares $\boldsymbol{x}$ and the synthetic sample made by $G$ and outputs a probability value based on the classification of $D$.
	%cGAN is a variation of GAN in which extra information such as a class label is additionally given as a condition. A generator $G$ and a discriminator $D$ are alternately trained to maximize the cost function in $D$'s side for discriminating original samples from the synthesized ones, and minimize in $G$'s side for generating realistic samples that are hardly distinguished from the original sample $\boldsymbol{x}$. The cost function of cGAN is the same as GAN except for the conditional probability distribution on the class label $\boldsymbol{c}$:	$\min_G \max_D V(D,G) = \mathbb{E}_{\boldsymbol{x} \sim p_{\textrm{data}}}(\boldsymbol{x}) [\log D(\boldsymbol{x}|\boldsymbol{c})] + \mathbb{E}_{\boldsymbol{z}\sim p_z} (\boldsymbol{z}) [\log (1-D(G(\boldsymbol{z}|\boldsymbol{c})))]$. 

	During the transmission, the participants want to keep their local data private against the server. With this end, each device selects dummy samples, whose label is not the target one, from its own dataset and enclose them when sending target-labeled seed samples. This operation hides the target labels from the server as it cannot discriminate between the received target and dummy labels.
	
	To address this label-wise privacy, we define the label privacy guarantee as the amount of label privacy preservation between interconnected nodes. Let $\boldsymbol{y}_i^T$ be a 1-dimensional logical vector whose elements are the target labels (private SDI) of $\textsf{device}_i$ that the server successfully receives. Similarly, $\boldsymbol{y}_i$ is a vector of labels including dummy labels (public SDI). Then, the label privacy of $\textsf{device}_i$ guaranteed from the server $\textrm{Priv}^{(1)}_i$ is defined as
	\begin{align} \label{eq:labelpriv_1hop}
		\textrm{Priv}^{(1)}_i = 1-|\boldsymbol{y}_i^T|/|\boldsymbol{y}_i|
	\end{align}
	where $|\cdot|$ indicates the sum of all elements of a vector. A device can keep its dataset information (represented as class label distribution) more private if it includes more dummy labels, while the label privacy guarantee is constrained by a ratio of target  to non-target labels. Note that if a device fails to upload part of its seed samples, the server does not receive the  label corresponding to the failed sample.  This incomplete  leads to a weaker guarantee for the device's label privacy due to the reduced  number of dummy labels.
	
	The overall latency in single-hop FAug is calculated as the transmission delay until the seed samples from all participants arrive at the server. In other words, the longest delay among all devices equals to the overall latency of the system. The overall uplink latency is measured as
	\begin{align} \label{eq:latency_1hop}
	L^{(1)} = \max_{i \in \mathcal{D}} \{T_i\}
	\end{align}
	Since the last arriving seed sample decides the overall latency, devices located far away are prone to path loss, which undermines the training step at the server.

	\section{Multi-hop Federated Augmentation With Sample Compression}

	\begin{figure*} [!t]
		\centering
		\includegraphics[width=\textwidth]{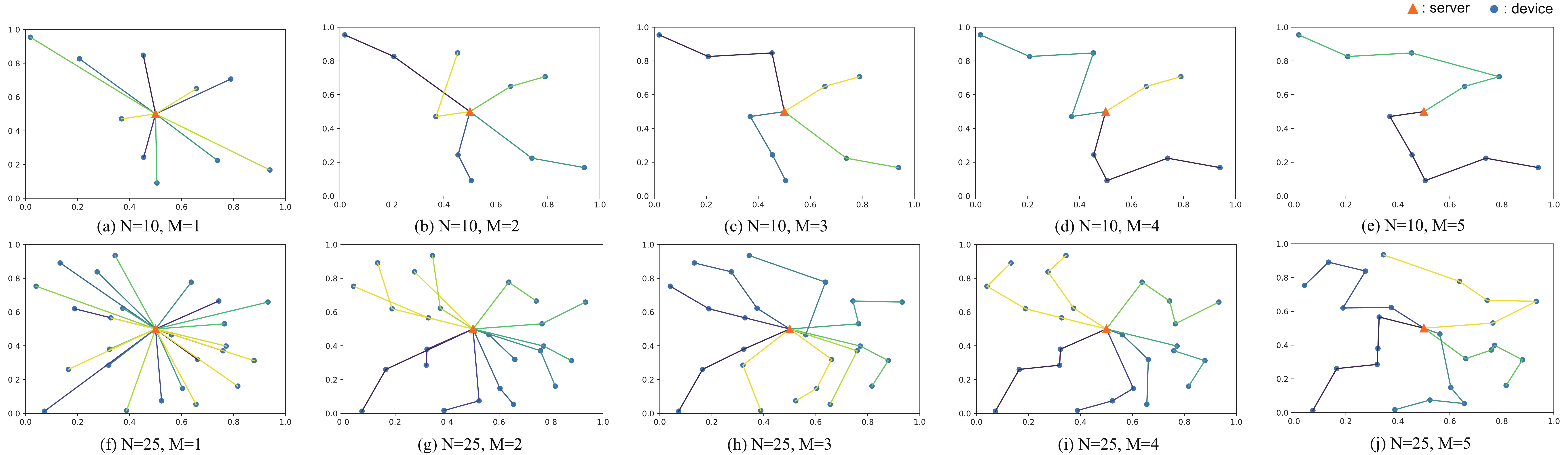}
		\caption{Exemplary topologies of single-hop and multi-hop scenarios with respect to the number of devices and maximum hops $(N=\{10,25\}, M=1,2,...,5)$}
		\label{fig_topology}
	\end{figure*}

	\begin{table}[t]
		\centering
		\scalebox{0.9}{
			\begin{tabular}{ll}
				\toprule   
				Variable & Description\\ 
				\midrule 
				$N$ & Number of devices ($|\mathcal{D}|=N$) \\
				$M$ & Max. number of hops in the system \\   
				$R_j$ & Set of devices transmitting through $j^{th}$ route\\
				& $(R_j \subset \mathcal{D})$\\ 
				$r$ & Number of total routes ($\max r=\lfloor\frac{N}{M}\rfloor$) \\
				$f^m(i)$ & $m$-hop destination of $\textsf{device}_i$  ($f^0(i)=i$)\\
				$\rho$ & Compression rate \\
				$Q_\rho$ & Compression mechanism with rate $\rho$\\
				$T_i$ & Latency\\
				$\tau$ & Deadline threshold ($T_i\leq\tau$ for $\forall i$)\\
				$\boldsymbol{y}_i^T$  & Target label indicator (private SDI) of $\textsf{device}_i$\\
				$\boldsymbol{y}_i^{dum}$  & Dummy label indicator of $\textsf{device}_i$\\
				$\boldsymbol{y}_i$  & Total label indicator (public SDI) of $\textsf{device}_i$\\
				& $(= \boldsymbol{y}_i^T \lor \boldsymbol{y}_i^{dum})$\\
				$\mathcal{X}_i$ & Set of data samples that $\textsf{device}_i$ wants to send\\
				$\mathcal{\hat{X}}_i$ & Set of compressed data samples in $\textsf{device}_i$\\
				$l$ & Label privacy threshold $(|\boldsymbol{y}^{dum}_i| < l)$\\				
				$b$ & Number of seed samples per label (upper bound)\\
				\bottomrule
			\end{tabular}
		}
		\caption{Notation and description in MultFAug} 
	\end{table}
	%\vspace{-10pt}

	%To overcome the path attenuation due to the significant distance from the server, the devices upload their seed samples through multiple hops.
	The single hop FAug has inherent problems in terms of privacy leakage for target labels and small uplink coverage. To tackle these issues, we propose a \emph{multi-hop federated augmentation with sample compression (MultFAug)} where devices construct a multi-hop route to the destination server and transmit their own compressed target samples.
	The proposed MultFAug hides sources of target samples from an aggregating server and intermediate devices, while guaranteeing the privacy leakage of individual delivered samples. Furthermore, this multi-hop feature extends the range of FAug by associating more devices to the server via multi-hop device-to-device communications, as elaborated in the following subsections.
	
	%The multi-hop protocol ensures that any intermediate device does not know where the samples arrived at it originally comes from~\cite{Liu18}.
	%Each device decreases the size of its data samples by sparsifying them before tossing to the destination. As a result, the sparsified samples of a device acquire privacy protection against other devices.

	\subsection{Multi-hop Protocol}

	We leverage multi-hop communications in order to increase coverage of the devices and provides higher privacy guarantee. 
	Multi-hop communications can reduce uplink latency by exploiting multiple successes of short path between devices. Especially, a device at the edge of the coverage stops suffering from sequential transmission outage if the multi-hop communications is applied. In addition, this multi-hop feature makes the server unable to discriminate the original source device of received each target labels because it receives only a mixed set of target labels and samples. This is consistent for intermediate devices. The proposed protocol of the multi-hop FAug is elaborated as follows.

	%When linking the devices into a single route, a device chooses its nearest neighbor preferentially. To avoid an endless stagnation of samples, the candidate neighbors that have already been selected as relaying nodes are excluded from the selection. A collection of maps with different number of devices and hops is exemplified in Figure~\ref{fig_topology}.
	
	Let us assume that the route from devices to the server is configured by the system. For a given routing path, a device sends its seed samples to another device as a relay node towards the server.
	%\tgray{We apply multi-hop communication to FAug that increases coverage of the devices and provides higher privacy guarantee. The key idea of communication efficiency due to multi-hop protocol is to reduce transmission delay by multiple success of short path communications. In addition, these relaying transmissions make each device unable to discriminate which label is whose target because it receives a mixed pack of seed samples from multiple devices. In a multi-hop protocol, a device sends its seed samples using another device as relay to reach the server.}
	Devices on the same route accumulate their samples over hops as described in Algorithm 1. As a result, devices located nearby the server have to carry more samples than those distant from the server. When a relaying node receives samples with public SDI, it attaches its own seed samples and overwrites their labels onto the public SDI. 
	In this research, the devices are interconnected with fixed number of hops.
	As in the single hop FAug, each device avoids sending more than $b$ samples that share the same labels. 
	If the target label of $\textsf{device}_i$ is already in the public SDI ($\boldsymbol{y}_i$), it discards random samples in its target label and replaces them from its local training set, so that any samples of the labels in $\boldsymbol{y}_i$ do not exceed $b$. 
	%The rules to determine the number of dummy labels are elaborated below.
	If an intermediate device receives an aggregated seed samples that already include its target label, the device does not add any dummy label at its transmission phase.

	Consider a function $f:\mathcal{D} \rightarrow \mathcal{D}$ that outputs an index of a device's direct destination through its route. Following this definition, the index of $\textsf{device}_i$'s direct destination is $f(i)$. Likewise, its $m$-hop destination is $f^m(i)$.
	The target labels of $\textsf{device}_i$ is label-private against its $M$-hop destination, and the label privacy guarantee is
	\begin{align} \label{eq:labelpriv_mhop}
	\textrm{Priv}^{(M)}_i = 1-\frac{|\boldsymbol{y}_i^T|}{\left|\bigvee\limits_{m=0}^M \boldsymbol{y}_{f^m(i)} \right| }
	\end{align}
	%where the function $f:\mathbb{N} \rightarrow \mathbb{N}$ indicates the routing algorithm that maps an index of a device onto that of the relaying device. 
	When the relaying transmissions succeed in a row, the seed sample set becomes larger as the seed samples are accumulated across the devices. Accordingly, total labels that the $M$-hop destination receives grows, i.e., $|\boldsymbol{y}_i| \leq |\boldsymbol{y}_i \vee \boldsymbol{y}_{f(i)}| \leq ... \leq |\vee_{m=0}^M \boldsymbol{y}_{f^m(i)}|$.
	This trend does not appear clearly when a device partially or fully fails to send its seed samples under tight communication constraints.
	%\tred{The denominator does not increase as $M$ increases.}

	The latency in MultFAug is denoted as $T_i$, which occurs between $\textsf{device}_i$ and its relaying device. None of the two devices in the same route occupy the bandwidth at the same time. The server waits for sample arrival from all routes. Assume that there are $r$ different routes and each of them binds at most $M$ devices delivering the seed samples through $M$-hop relays. The latency in a route is calculated as the sum of latency at each hop. The overall latency in the system using a multi-hop protocol is	
	\begin{align}
	\label{eq:latency_Mhop}
	L^{(M)} = \max_{j \in \{ 1,2,...,r \}}  \sum_{i \in R_j}{T_i} 	
	\end{align}
	where $R_j$ indicates the $j\textsuperscript{th}$ route. The overall latency in (\ref{eq:latency_1hop}) is thereby interpreted as a special case in which $M=1$. In single hop protocol, the number of routes is the same as that of devices because each device independently has its own route to the server. That is, all $T_{R_j}=\sum_{i \in R_j}{T_i}$ is identical to $T_i$.
	
	\begin{algorithm}[!t]
		\caption{\textsf{Device}$_i$'s sample compression and transmission for target and dummy labels in MultFAug}
		\begin{algorithmic}[1]
			%\renewcommand{\algorithmicrequire}{\textbf{Input:}}
			%\renewcommand{\algorithmicensure}{\textbf{Output:}}
			%\small 
			\REQUIRE $R_j=\{i|i=1,2,\dots,M\}$ $(j=1,...,r)$ where $i=1$ is the edge device of $j^{th}$ route and $f(i)=i+1$ $\forall i$, label privacy threshold $l$, compression rate $\rho$
			\FOR{$j=1$ to $r$}
			\WHILE{$i<M$}	    
			\IF{$d$}
			\STATE $\mathcal{X}_i \gets \mathcal{X}^T_i \cup \mathcal{X}^{dum}_i$
			\STATE $\mathcal{\hat{X}}_i \gets Q_\rho(\mathcal{X}_i)$ \\
			(Compress $\mathcal{X}_i$)
			\STATE $\boldsymbol{y}_i \gets \boldsymbol{y}^T_i \lor \boldsymbol{y}^{dum}_i$
			\ELSE  % i is a relaying node
			\STATE Receive $\mathcal{X}_{i-1}$, $\boldsymbol{y}_{i-1}$
			\IF{$|\boldsymbol{y}_{i-1} \lor \boldsymbol{y}_i^T| \leq l$}
			\STATE $\mathcal{X}_i$, $\boldsymbol{y}_i \gets \mathcal{X}_{i-1}$, $\boldsymbol{y}_{i-1}$
			\ELSE
			\STATE Set $\boldsymbol{y}_i^{dum}$ such that $|\boldsymbol{y}_i^{dum}| \gets l - |\boldsymbol{y}_{i-1} \lor \boldsymbol{y}_i^T|$
			\STATE $\mathcal{X}_i \gets \mathcal{X}^T_i \cup \mathcal{X}^{dum}_i$
			\STATE $\mathcal{\hat{X}}_i \gets Q_\rho(\mathcal{X}_i)$ 
			\STATE $\boldsymbol{y}_i \gets \boldsymbol{y}^T_i \lor \boldsymbol{y}^{dum}_i$
			\ENDIF		  
			\ENDIF
			\STATE Transmit $\mathcal{X}_i$ to $\textsf{device}_{i+1}$
			\ENDWHILE
			\STATE Transmit $\mathcal{X}_M$ to the server
			\ENDFOR
		\end{algorithmic}
	\end{algorithm}

	\begin{algorithm}[!t]
		\caption{\textsf{Server}'s sample collection, oversampling, and cGAN training.}
		\begin{algorithmic}[1]
			%\small
			\REQUIRE deadline threshold $\tau$, route $R_j (j=1,2,...,r)$
			\FOR{$t=1$ to $\tau$}
			\FORALL{$R_j$}
			\IF{$\mathcal{X}_j$ arrived through $R_j$}
			\STATE samples $\gets$ samples $\cup$ $\mathcal{X}_j$
			\STATE $\boldsymbol{y} \gets \boldsymbol{y} \lor \boldsymbol{y}_j$
			\ENDIF
			\ENDFOR
			\ENDFOR
			\STATE real samples $\mathcal{X}_{real} \gets$ Oversample data with $\boldsymbol{y}$
			\WHILE{$G$'s parameters have not converged}
			\STATE Select a random batch from the real samples
			\STATE Generate noise $\boldsymbol{z} \sim p(z)$
			\STATE Generate synthetic samples $\mathcal{X}_{gen} \gets G(\boldsymbol{z},\boldsymbol{c})$
			\STATE Calculate loss with real samples \\
			$D_{loss}=\log D(\mathcal{X}_{real} | \boldsymbol{c})$
			%\STATE $D_{loss} \gets 0.5*(D_{loss}^{real} + D_{loss}^{gen})$
			\STATE $\boldsymbol{c}\textsubscript{SDI} \gets \boldsymbol{y}$
			\STATE Calculate loss with synthetic samples \\
			$G_{loss} \gets \log (1-D(G(\boldsymbol{z} | \boldsymbol{c}\textsubscript{SDI})))$
			\ENDWHILE
			\RETURN $G$
			\FORALL{$i$}
			\STATE Transmit $G$ to $i$
			\ENDFOR
		\end{algorithmic}
	\end{algorithm}
	%\vspace{-20pt}
	
	\subsection{Sample Compression}

	We include a compression mechanism in the proposed protocol in order to reduce the communication overhead and privacy leakage of each seed sample. Specifically, each device compresses its own sample to be delivered by discarding randomly selected bits from the sample. The number of the selected bits is determined by a compression rate $\rho$, which is decided a priori.
	Note that the similarity between two compressed samples becomes larger compared to that between the original ones. This indicates that the sample compression makes it more difficult to distinguish those samples, leading to stronger local privacy preservation.

	For both sample privacy and communication efficiency, each device discards some of the bits from the data samples. The number of removing bits depends on the compression rate $\rho$. The sample distortion also results in obfuscation between two different images as the image similarity increases. The difficulty in distinguishing those samples leads to stronger local privacy preservation.
	
	Specifically, the proposed sample compression consists of sparcification and transformation to a compressed sparse row matrix.
	At first, we provide a randomized sparsification algorithm denoted by $Q_\rho:X \rightarrow \hat{X}$ where $X$ and $\hat{X}$ indicates the set of bits composing the original sample and its sparsified version, respectively.
	Let us consider that the devices train with the MNIST dataset in the MultFAug system. Among $28\times28$ elements of a handwritten image, $\lfloor \rho \times28\times28 \rfloor$ arbitrary elements are chosen and converted to zero.
	After converting partial elements into zero with rate $\rho$, the device compresses each of the distorted sample into a sparse matrix, resulting in smaller data size. 
	Specifically, the devices transmit their seed samples as a form of compressed sparse row (CSR) matrices that consist of the original matrices' nonzero elements, list of row and column indices for the nonzero elements \cite{Saad03}. Hence if a distorted sample contains a number of zero values, its sparse matrix saves on communication payload.
	The data size of $\hat{x}$ averages out at 1/5 of the original data size, although it varies according to the number of nonzero elements in each data sample. The reduction of data size lessons communication payload and therefore achieves less latency. %\tblue{Elaboration: CSR format}

	When devices transmit their data samples to the relaying device, they disclose their local information. In case of transmitting through multiple hops, the local information inevitably reaches the server. Therefore, a device locally distorts its data samples before uploading it to restrict the leakage of local information.

	Local differential privacy (local DP) is provided in contexts where each participant distorts its local data samples to keep them private before they reach to other entities~\cite{Kairouz14}. While samples guaranteed local DP need distortion at every element of each sample, our sample compression removes partial elements of a sample to increase the similarity of the samples in the dataset. Therefore, we suggest a measure of sample privacy guarantee inspired from local DP.
	Local differential privacy implies that the output reveals the limited information of the input. If any data sample $x, \tilde{x}$ in the original dataset $\mathcal{X}$ transform into identical output $\hat{x}$ among the codomain $\mathcal{\hat{X}}$ with higher probability, it infers that the mechanism privatizes each sample from the others more strictly~\cite{Xiong16}.
	For a non-negative $\epsilon$, a privatization mechanism $Q$ is \textit{$\epsilon$-locally differentially private} if $\max_{(x,\tilde{x},\hat{x}) \subset \mathcal{X} \times \mathcal{X} \times \mathcal{\hat{X}}} \big\{ Q(\hat{x}|x)/Q(\hat{x}|\tilde{x}) \big\} \leq e^\epsilon$. 
	In similar fashion, our randomized compression mechanism $Q$ yields high image similarity between any two different data samples: for any $x, \tilde{x}$ in the dataset, the distance between their output $Q_\rho(x)$ and $Q_\rho(\tilde{x})$ is short.
	
	%For any other input $\tilde{x}$, the output under $\tilde{x}$ is sufficiently similar to that under $x$.
	%The sample privacy implies that the sparsified output under any other $\tilde{X}$ has a short distance to that under $X$. If the sparsified version of two different images $X$ and $\tilde{X}$ ($Q(X)$ and $Q(\tilde{X})$) have closer distance than $\delta$ in multidimensional space, they are considered to be identical.
	%\tgreen{ We are interested in locally differentially private requantization: we would like to find a randomized mapping $Q:\mathcal{X} \rightarrow \hat{\mathcal{X}}$ where $\hat{\mathcal{X}}\subset\mathbb{R}$ is an output discrete set of points and $|\hat{\mathcal{X}}| \leq |\mathcal{X}|$. This requantization mapping can be thought of as a channel $Q(\hat{x}|x)$ (conditional probability distribution). Our goal is to pass $X_n$ through this channel $Q$ to release a privatized and compressed version $\hat{X}_n$. A channel $Q$ is $\epsilon$-locally differentially private [9, 10] if}

	The sample privacy guarantee of the dataset is defined as the inverse of image similarity. 
	Without loss of generality, the image similarity of data samples in the accumulated dataset $\mathcal{X}$ is measured as
	\begin{equation} \label{eq:sample priv}
	\textrm{sim}(\mathcal{X}) = \max_{(x,\tilde{x}) \subset \mathcal{X} \times \mathcal{X}} \log d(Q_\rho(x), Q_\rho(\tilde{x}))
	\end{equation}
	for any $x, \tilde{x}$.
	%Small similarity implies tighter bound for the sample privacy guarantee in a network as it is the log-distance between any two data samples.
	To measure the similarity in (\ref{eq:sample priv}), we apply classical multidimensional scaling (MDS) algorithm, a statistical technique used to represent each image by a point in a lower dimensional space. To find the configuration in the low-dimensional space, the loss function in MDS called \emph{stress} should be minimized. It is known that the direct solution for the classic MDS problem exists when the pointwise distances of converted samples are calculated as Euclidean distance.
	The configurated map displays the optimal distances among the converted points, which visualizes the level of similarity~\cite{Mair18,Hout16}. 
	
	%Multidimensional scaling is the problem of representing each image $I_i \in ohm$ by a point (vector) in a low dimensional space $x_i \in R$

	%Both in single hop and protocol, the overall latency depends on the last arriving sample.

	\section{Numerical Results}
	
	\begin{figure*} [!t]
		\centering
		\subfigure[]{\includegraphics[width=0.32\textwidth]{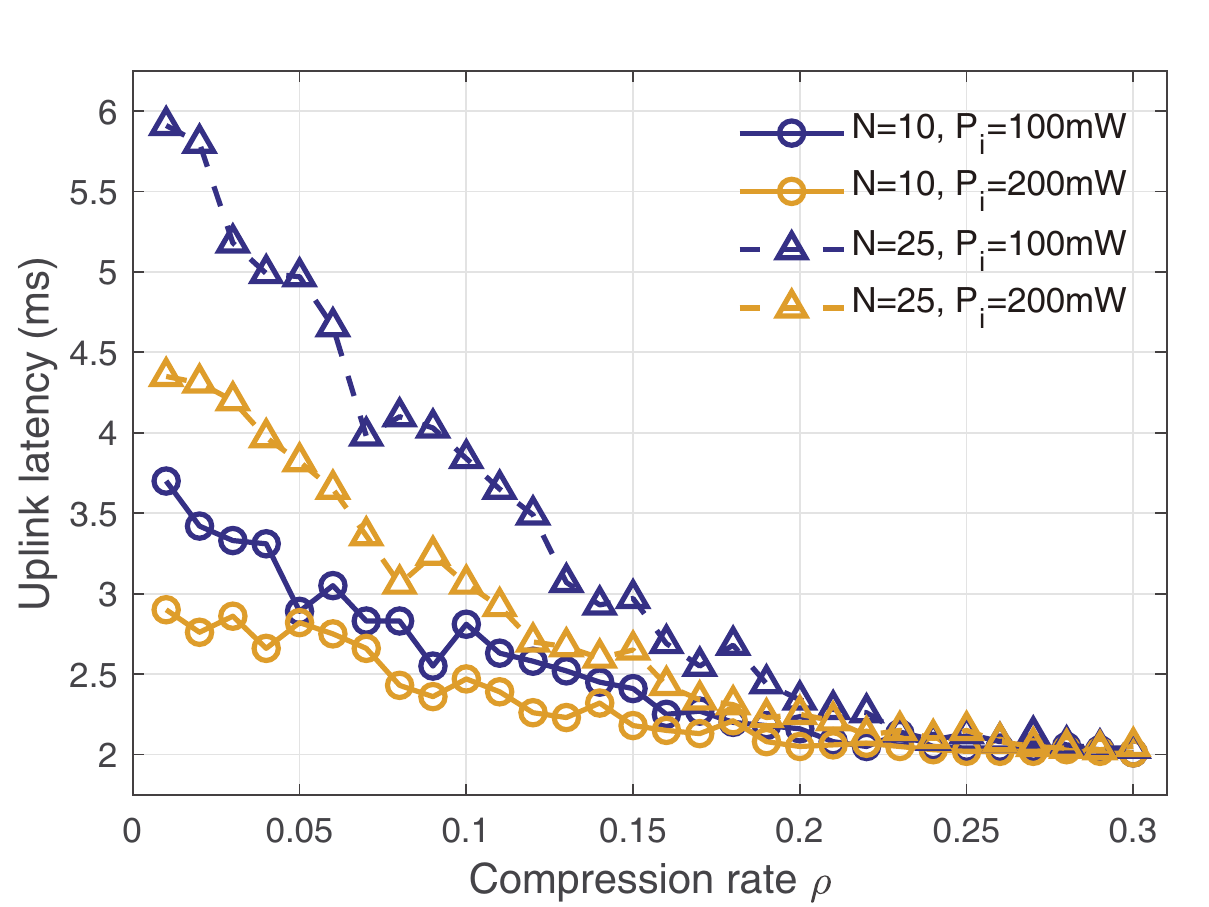} \label{fig3a}}
		\subfigure[]{\includegraphics[width=0.32\textwidth]{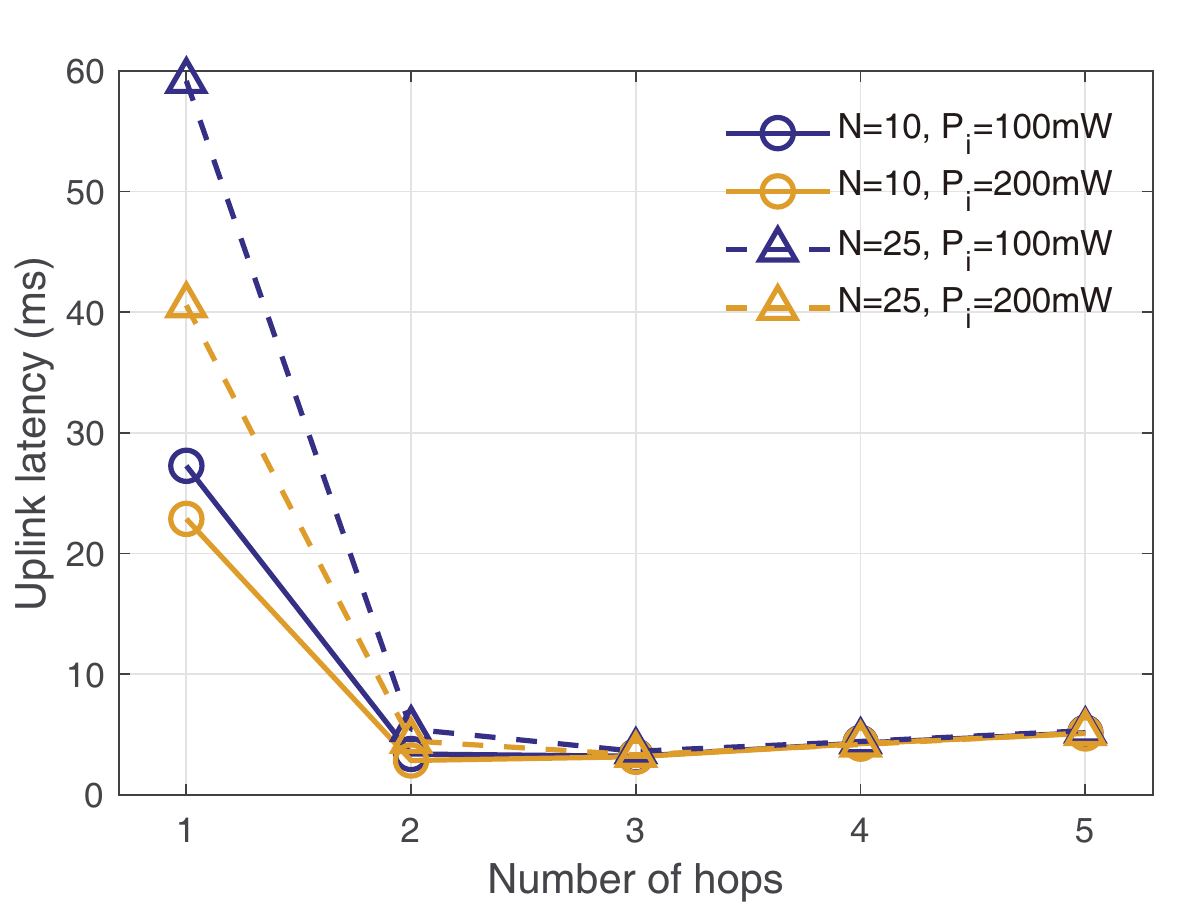} \label{fig3c}}
		\subfigure[]{\includegraphics[width=0.32\textwidth]{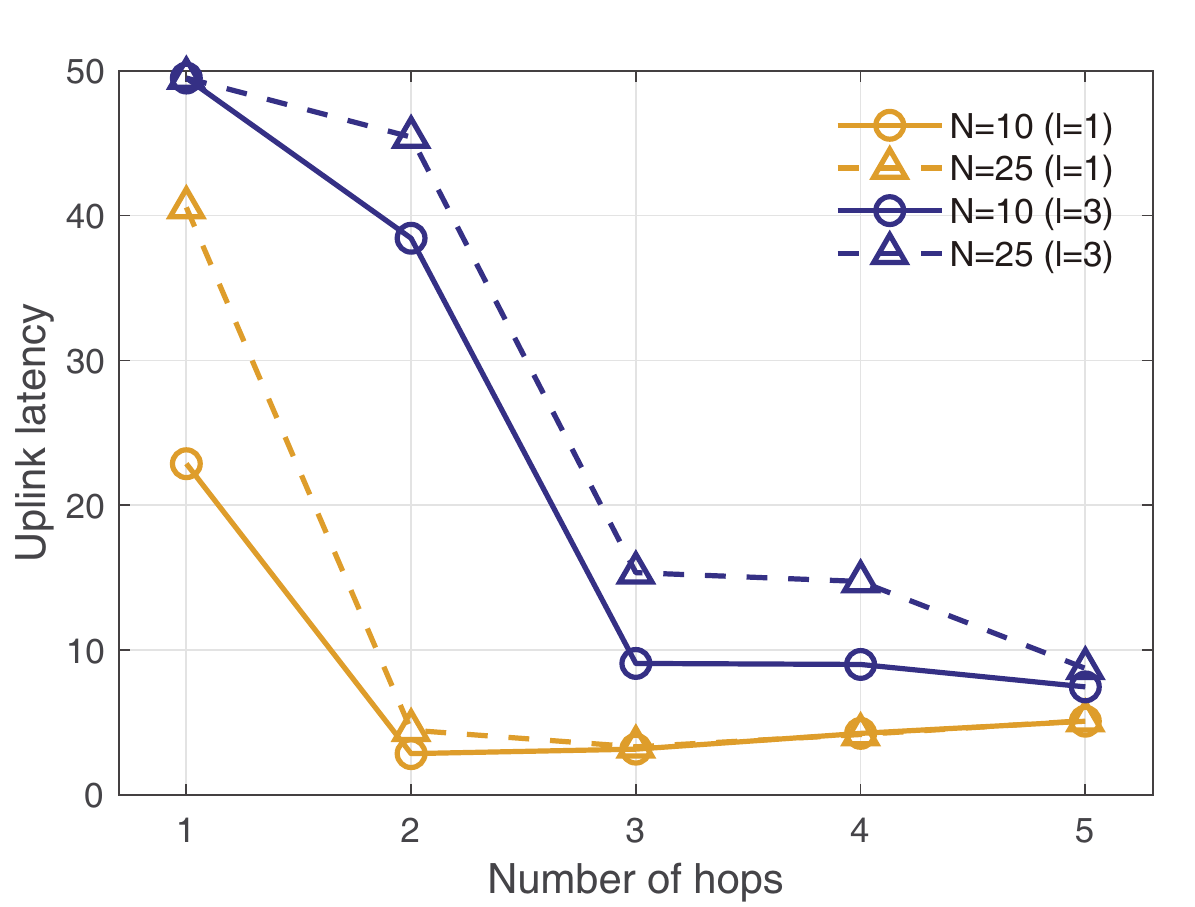} \label{fig3d}}
		\caption{Uplink latency with respect to (a) compression rate $\rho$ with different transmission powers, (b) the number of hops with different transmission powers, (c) the number of hops with different label privacy thresholds $l$ ($\rho=0.02$). }
		\label{neofig3}
	\end{figure*}	
	
	\begin{figure*} [!t]
		\centering
		\includegraphics[width=\textwidth]{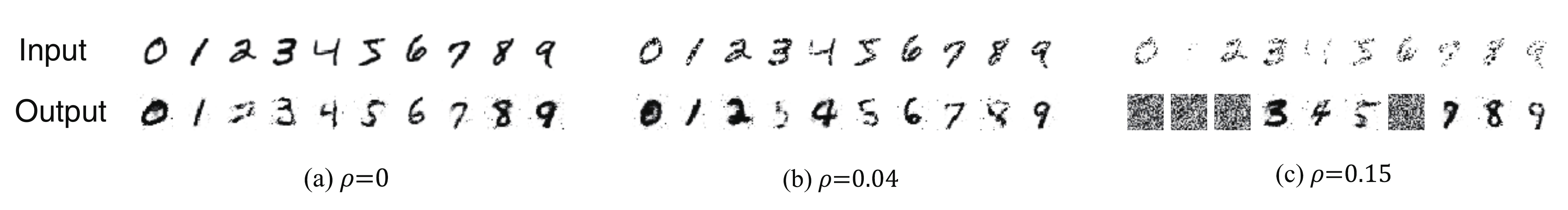}
		\caption{Input and output samples of the generator, which is a part of the trained cGAN in the server for different compression ratios: (a) $\rho=0$, (b) $\rho=0.04$, and (c) $\rho=0.15$.}
		\label{fig4}
	\end{figure*}

	In this section, we numerically evaluate the performance of the proposed MultFAug, in terms of its communication efficiency and privacy guarantee. The network topologies under study is visualized in Figure~\ref{fig_topology}.
	
	\subsection{Experimental Settings}
	
	Namely, we consider $N=\{10,25\}$ devices that are uniformly distributed over a two-dimensional network plane with the size $10 \times 10 \textrm{km}^2$. From a single server located at the center of the plane, multiple device-to-server routes are constructed in a nearest-neighbor rule, in a way that each route comprises up to $M\leq 5$ hops. Every device therein has 1,804 MNIST samples with $10$ labels, comprising $4$~samples in a single target label and $200$ samples for each non-target label. Each device's target label is uniformly randomly selected, leading to the non-IID MNIST dataset dispersed across devices. 
	%The label privacy of each device is bounded to at least 0.5 by adding one dummy label to the edge devices. An edge device with one target label has an private SDI with only one nonzero element.
	The edge devices can bound the minimum amount of their label privacy guarantee by adding $l$ dummy label(s). For instance, setting $l=1$ by default brings about $1-(|\boldsymbol{y}^T_i|/|\boldsymbol{y}_i|)=1/2$ for any $i \in \mathcal{D}$. This implies that the label privacy of each device is bounded to 0.5 in direct transmission between any two devices.
	The communication and privacy performance is evaluated based on a reference device, the farthest device from the server within a route that is uniformly randomly selected. The reference device represents the tendency of all devices by averaging the results with sufficient amount of iterations.
	
	For MultFAug, the server trains a cGAN~\cite{Mirza14} consisting of a 4-layer discriminator network, as well as a 4-layer generator network with 1,493,520 weight parameters. During local training after MultFAug, following~\cite{Jeong18}, each device utilizes a convolutional neural network with 1,199,776 weight parameters, consisting of $2$ convolutional layers, $1$ max-pooling layer, $1$ flattened layer, and $2$ fully connected layers. Other default simulation settings are summarized as follows: $\alpha=4$, $B=20\textrm{MHz}$, $P_i=200\textrm{mW}$, $N_0=-174\textrm{dBm/Hz}$, $h^2 \sim exp(1)$.
	
	%In Figure \ref{fig3c} and Figure \ref{fig3a}, only the transmission delay in the uplink communication is considered for the uplink latency. Figure \ref{fig3a} shows the results for a 2-hop protocol. Figure \ref{fig5} represents the case where 10 devices participate in FAug under a single hop transmission. In Figure \ref{fig6} the compression rate is fixed to $\rho=0.02$.

	\subsection{Uplink Latency}

	Figure \ref{fig3a} illustrates that more sample compression ratio $\rho$ tends to decrease the reference device's uplink latency, thanks to the communication payload size reduction. The latency has a minor fluctuation over $\rho$, since lower $\rho$ does not always yield smaller CSR format data sizes (see the details in Sec.~III-B). The latency under $N=25$ is higher than that of $N=10$ due to its more accumulated volume of seed samples, as also observed in Figure~\ref{fig3c}. Figure \ref{fig3c} shows that the uplink latency from the reference device to the server is minimized at 2 hops for $N=10$ or 3 hops for $N=25$. The latency is convex-shaped with respect to the number $M$ of hops due to the following two conflicting behaviors. On the one hand, as $M$ increases, path losses are compensated thanks to the per-hop distance reduction. On the other hand, more hops are likely to incur more channel outages, while increasing the accumulated volume of seed samples from the preceding hops. Figure~\ref{fig3d} demonstrates the influence of minimum number of dummy labels on the overall latency. The latency is higher in a system that provides stronger label privacy in every direct transmission because of the increased number of seed samples to transmit.
	
	\subsection{Generator Performance}

	\begin{figure} [!t]
		\centering
		\includegraphics[width= 0.42\textwidth]{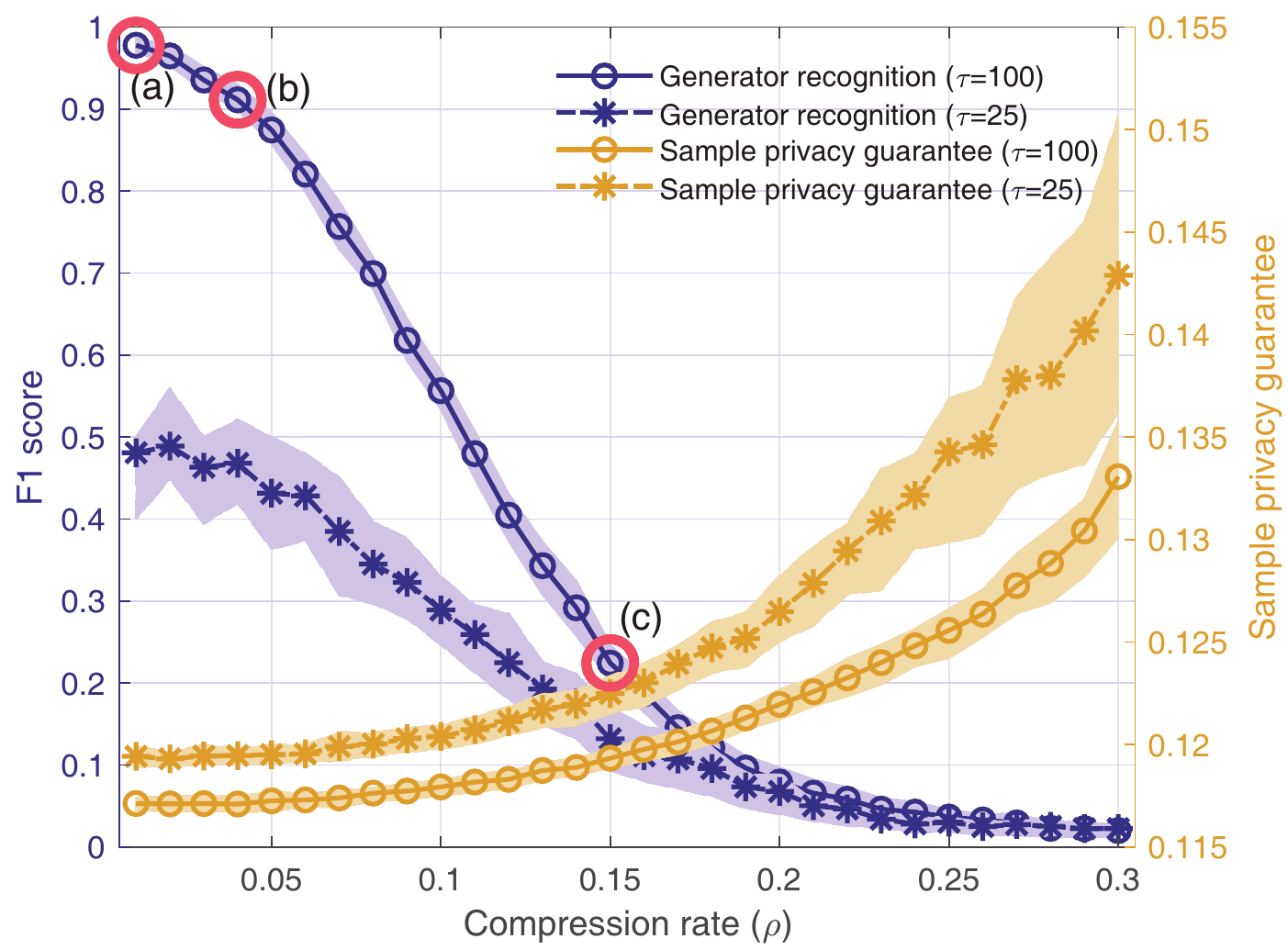}
		\caption{Trained generator's F1 score (left, violet) and sample privacy guarantee (right, yellow) with respect to sample compression rate $\rho$ ($N=10, M=2$).}
		\label{fig5}
		\vspace{-10pt} 
	\end{figure}

	\begin{figure*} [!t]
		\centering
		\subfigure[Test accuracy with latency deadlines ($\tau$).]{\includegraphics[width=0.32\textwidth]{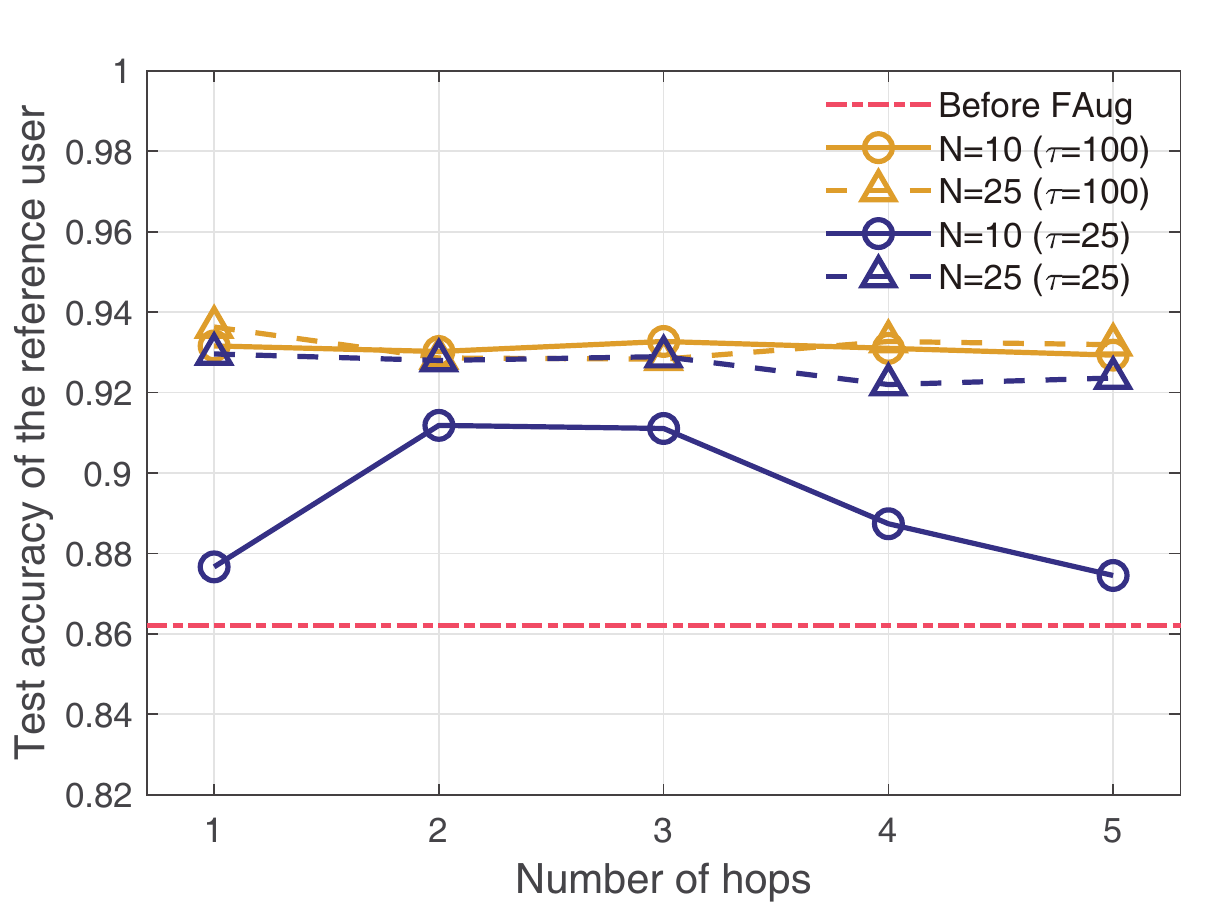} \label{fig6a}}
		\subfigure[Test accuracy with label privacy guarantee deadlines ($l$).]{\includegraphics[width=0.32\textwidth]{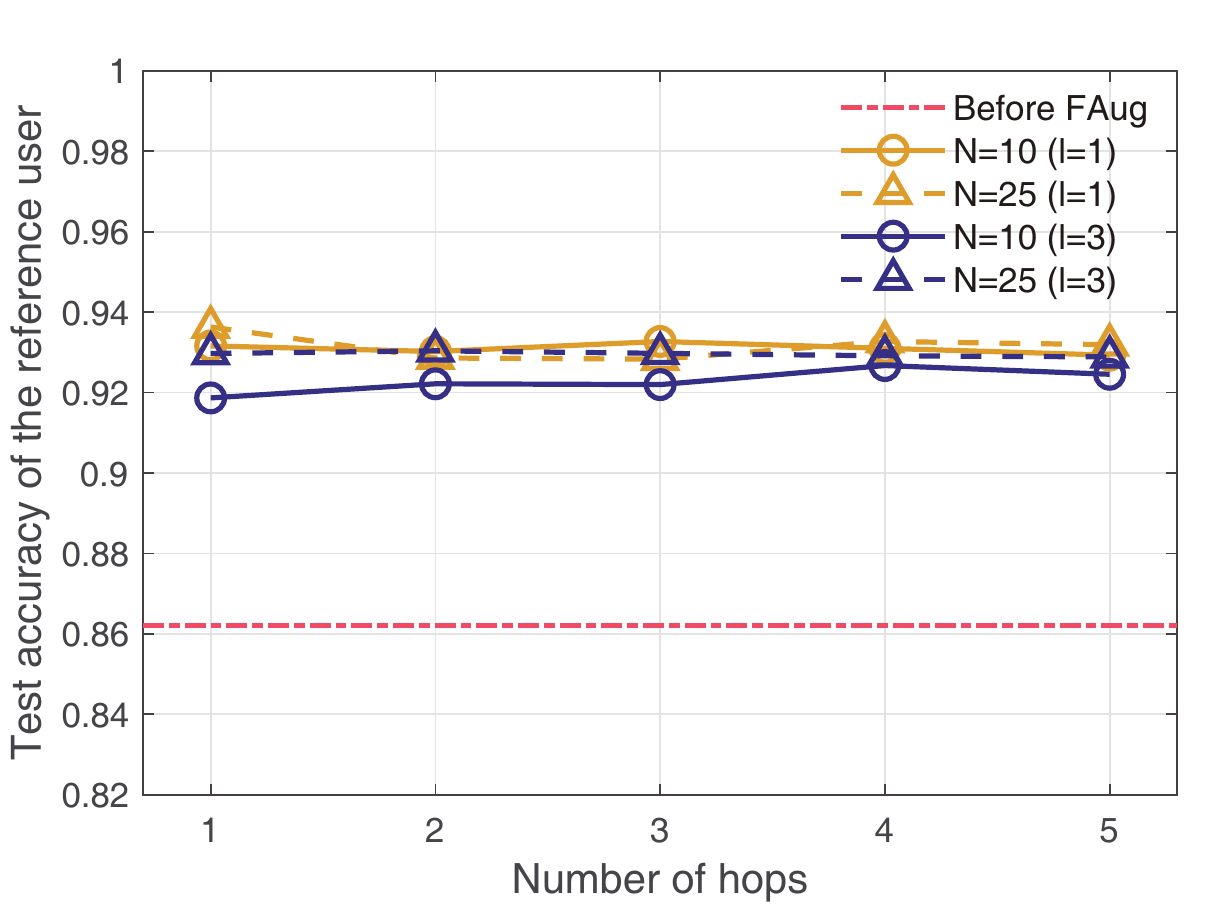} \label{fig6b}}
		\subfigure[Label privacy guarantee with latency deadlines.]{\includegraphics[width=0.32\textwidth]{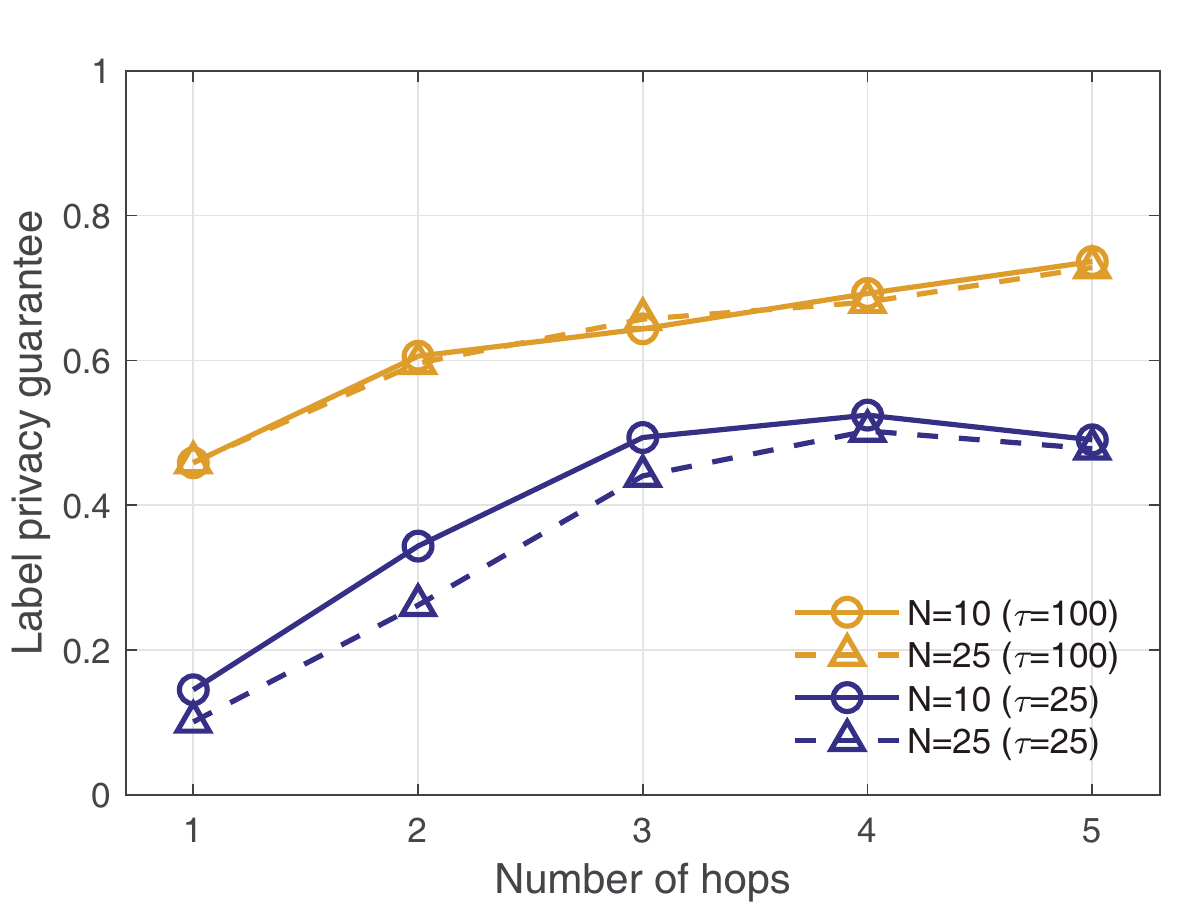} \label{fig6d}} 
		\caption{Evaluations with respect to the number of hops $M$. Test accuracy of the reference user after operating FAug in single hop and multi-hop protocols with different (a) latency thresholds, (b) label privacy guarantee thresholds, (c) label privacy guarantee with different latency thresholds ($\rho=0.02$).} 
		\label{neofig6}
	\end{figure*}
	
	For different $\rho$'s, Figure \ref{fig4} visualizes the aggregated seed samples at the server, i.e., the input samples for cGAN training. Each output sample of the trained cGAN's generator is obtained by feeding a Gaussian noise and a class label. With $\rho=\{0, 0.04\}$, the generator can successfully produce augmented digits. On the contrary, with $\rho=0.15$, the generator fails to yield augmented samples for the digits 0, 1, 2, and 6, highlighting the importance of optimizing $\rho$.

	Figure \ref{fig5} shows the performance of the trained generator (left, violet) and sample privacy guarantee (right, yellow) with respect to $\rho$. The generator's performance is measured using the F1 score that is the harmonic mean of precision and recall, where precision increases with the similarity between the generated and ground-truth images, and recall increases with the number of generable images. As $\rho$ increases, while the sample privacy guarantee increases, the cGAN is trained with more noisy samples, degrading the trained generator's F1 score. With a more stringent sample collection latency deadline $\tau$, the server collects fewer seed samples, leading to the generator's lower F1 score while increasing the sample privacy guarantee.

	\subsection{Test Accuracy of Local Models After MultFAug}

	Lastly, Figure \ref{fig6a} and \ref{fig6b} illustrate the reference device's test accuracy before and after applying MultFAug with respect to the number of hops. With a long collection latency deadline $\tau$ and/or a large number of devices, the server can collect a sufficiently large number of seed samples for training the generator of MultFAug. The plenty of seed samples also improve the reference device's local training after applying MultFAug. Therefore, the abundance of acquired seed samples enables the device to achieve high test accuracy, regardless of the communication protocol characterized by the number of hops. On the contrary, with a more stringent latency deadline $\tau=25$, the test accuracy is maximized at around 2-3 hops that provide the minimum uplink latency as observed in Figure~\ref{fig3c}. 
	When the deadline is tighter, the number of hops weighs heavier with the range of the amount of arrived seed samples. Therefore, the inverse tendency is shown in uplink latency and test accuracy.
	
	In contrast, label privacy guarantee increases with the number of hops, as shown by Figure~\ref{fig6d}. Note that the larger number of hops does not always ensure the stronger label privacy, specifically in the cases under tighter time deadlines. For instance, when the latency deadline is $\tau=25$, the server is less likely to receive all seed samples in time. The multi-hop protocol makes each sample to pass through more devices, although it has an advantage of shortening the travel distance of each sample per one transmission. As a result, devices with a tighter latency deadline show concave curves with an optimal $M=4$.
	 
	%\tblue{why in Fig. 6(c) there is an optimal M=4 in terms of privacy for the cases of r=25?}

	%More compression $\rightarrow$ relieves communication payload $\rightarrow$ takes faster to reach the server.	
	%Sparsification on every sample requires smaller communication cost.

	% Devices sharing the same route are scheduled not to try sending data samples at the same time. Since the number of active channels decreases as we combine more devices in each route, the multi-hop system can allocate more bandwidth resource to each channel. Those channels with larger transport capacity results in lower total uplink latency.

	\section{Conclusion}
	
	In this paper, we proposed a multi-hop federated augmentation with sample compression (MultFAug) scheme. MultFAug allows devices to augment data samples by hiding their target labels in multi-hop communications while reducing the device-to-server uplink latency. In addition, the sample compression of MultFAug preserves the privacy of each seed sample, and reduces the communication overhead at the same time. The effectiveness of MultFAug was validated by numerical evaluations, highlighting the importance of the number of hops and compression rate. For a given network topology and channel condition, jointly optimizing the number of hops and compression rate could thus be an interesting topic for future work.
	
	%\begin{figure} [!t]
	%	\centering
	%	\includegraphics[width=\columnwidth]{Figs/mixup_07.eps}
	%	\caption{some caption}
	%\end{figure}
	
	\section*{Acknowledgements}
	This research was supported partly by Basic Science Research Program through the National Research Foundation of Korea(NRF) funded by the Ministry of Science and ICT(NRF-2017R1A2A2A05069810), partly by Academy of Finland projects SMARTER, CARMA, and 6Genesis Flagship (grant no. 318927), and partly by AIMS and ELLIS projects at the University of Oulu.

	\bibliographystyle{named}  % appearance order
	\bibliography{MYBIB}
	
\end{document}